\documentclass{article}

\usepackage[english]{babel}
\pdfoutput=1

\usepackage[letterpaper,top=2cm,bottom=2cm,left=3cm,right=3cm,marginparwidth=1.75cm]{geometry}

\usepackage{amsmath}
\usepackage{graphicx}
\usepackage[colorlinks=true, allcolors=blue]{hyperref}
\usepackage{caption}
\usepackage{authblk}
\usepackage{subcaption}
\usepackage{float}

\date{}

\title{Optimizing Bayesian acquisition functions in Gaussian Processes}

\author[1]{Mr. Ashish Anil Pawar}
\author[2]{Dr. Ujwal Warbhe}
\affil[1]{College of Engineering, Pune}
\affil[2]{National Institute of Technology, Srinagar}
\affil[1]{pawaraa16.it@coep.ac.in}
\affil[2]{ujwalwarbhe@nitsri.net}

\begin{document}
\maketitle
\begin{abstract}
Bayesian Optimization is an effective method for searching the global maxima of an objective function especially if the function is unknown. The process comprises of using a surrogate function and choosing an acquisition function followed by optimizing the acquisition function to find the next sampling point. This paper analyzes different acquistion functions like Maximum Probability of Improvement and Expected Improvement and various optimizers like L-BFGS and TNC to optimize the acquisitions functions for finding the next sampling point. Along with the analysis of time taken, the paper also shows the importance of position of initial samples chosen.

\end{abstract}

\section{Introduction}

Bayesian optimization is a popular optimization technique for optimizing a black box function especially with high dimensions. For a known objective functions, various optimization functions are readily available to choose from. For a black box function, since the true nature of the objective function is unknown, many available optimization techniques including Gradient Descent cannot be applied.

For a black box function, various other optimization techniques are available such as Grid Search and Random Search, however, both of these techniques are extremely inefficient and time consuming specially if the objective function is costly to execute. Instead, Bayesian optimization tries to find the global optimum by using a surrogate function to evaluate the real objective function, thus, making the computation much efficient with respect to time or money.

\section{Background}

\subsection{Bayesian optimization}

Bayesian optimization aims to solve global optimization conundrum\cite{mockus1978application}. Consider a black box function $f: X \rightarrow Y$. Bayesian optimization tries to find the solution to the problem:
\begin{equation}
\max_{x \in A} f(x)
\end{equation}
along with minimizing the cost incurred while running iterations on the objective function $f(x)$. Bayesian optimization uses the popular Bayes' theorem which states that the posterior probability of a model $M$ given evidence $E$ is directly proportional to the product of likelihood of $E$ given $M$ and the prior probability of $M$:
\begin{equation}
P(M|E) \propto P(E|M) P(M)
\end{equation}
For the observed data
\begin{equation}
D_{1:t} = \{(x_i, y_i), i = 1, 2, \dots t\}
\end{equation}
the posterior distribution obtained for Bayesian optimization is:
\begin{equation}
P(f|D_{1:t}) \propto P(D_{1:t}|f) P(f)
\end{equation}
Bayesian optimization incorporates prior belief about $f$ and updates the prior with samples drawn from $f$ to get a posterior that better approximates $f$. A surrogate model is used for approximating the objective function. It also uses an acquisition function to help find the next sampling point where next best observation is likely to be found.
Bayesian optimization has found its applications in various fields like Material Science \cite{frazier2016bayesian} \cite{packwood2017bayesian} and parameter tuning \cite{falkner2018bohb}. Snoek's work \cite{snoek2012practical} opened the doors to new machine learning optimizations using Bayesian Optimization. Later Sweresky \cite{swersky2013multi} used Bayesian Optimization for multi-task optimization.

\subsection{Surrogate model}

Bayesian optimization uses a surrogate model to approximate the acquisition function by providing the probabilistic interpretation of $f$ over a finite set of points. The most commonly used surrogate model is Gaussian Process (GP). \cite{rasmussen2006cki} Gaussian Process is a collection of random variables so that the join distribution of every finite subset of random variables is a multivariate Gaussian:
\begin{equation}
f \sim GP(\mu, k),    
\end{equation}
where $\mu$ is the mean function and $k$ is a positive definite kernel function also called as covariance function.
For this research we use the Matérn 5/2 Kernel as it does not have the concentration of measure problems for high dimensional spaces:
\begin{equation}
K_{\text{M52}}(x,x') = \theta_0(1 + \sqrt{5r^2(x,x')} + \frac{5}{3}r^2(x, x'))exp\{-\sqrt{5r^2(x,x')}\}
\end{equation}

\subsection{Acquisition functions}

Gaussian Process utilizes acquisition functions for making decisions on where to sample. We assume that the function $f(x)$ is drawn from a Gaussian process prior and that our observations are of the form $\{X_n,y_n\}^N_{\text{n=1}}$, where $y_n \sim N(f(X_n),v)$ and $v$ is the variance of noise introduced into the function observations.

There are various acquisition functions like Maximum Probability of Improvement(MPI), Expected Improvement(EI), Upper Confidence Bound (UCB), Simple Regret(SR) and Entropy Search(ES). However the most commonly used acquisition functions are Probability of Improvement and Expected Improvement and hence the paper revolves around various optimization techniques for these acquisition functions and their impact on convergence to global optimum for the objective function.

\subsubsection{Maximum Probability of Improvement}
Maximum probability of improvement is the maximum probability of the surrogate function. The next query point having the maximum probability of improvement is given by:
\begin{equation}
x_{\text{t+1}} = argmax(P(f(x) \geq (f(x^+) + \varepsilon))) = argmax_x\Phi(\frac{\mu_t(x) - f(x^+) - \varepsilon}{\sigma_t(x)} )
\end{equation}
where,\\
$P(\cdot)$ indicates probability\\
$\varepsilon$ is a small positive number\\
and, $x^+ = argmax_{{x_i} \in x_{\text{1:t}}}f(x_i)$ where $x_i$ is the location queried at the $i^{th}$ step\\
$\Phi(\cdot)$ indicates the CDF

\subsubsection{Expected Improvement}
Expected Improvement is based on the simple idea to choose the next point to explore so that it has the highest improvement expected over the current max. Expected improvement is basically the trade off between exploration and exploitation. Expected improvement is given by the equation:
\begin{equation}
EI(x) = \begin{cases}
(\mu_t(x) - f(x) - \varepsilon)\Phi(Z) + \sigma_t(x)\phi(Z), & \text{if} \  \sigma_t(x) > 0\\
0, & \text{if} \ \sigma_t(x) = 0
\end{cases}
\end{equation}

\begin{equation}
Z = \frac{\mu_t(x) - f(x^+) - \varepsilon}{\sigma_t(x)}
\end{equation}
where,\\
$\Phi(\cdot)$ indicates CDF\\
and, $\phi(\cdot)$ indicates PDF

The algorithm was first cited in \cite{movckus1975bayesian} and later gained more popularity in \cite{jones1998efficient}.

\section{Experiments}
This paper focuses on the study of various global optimizers on acquisition functions also called as "inner optimization" in Bayesian optimization.

The experiments were conducted comparing two optimizers on Maximum Probability of Improvement and Expected Improvement acquisition functions.

\subsection{Objective function}
The experiments were carried out on the below multimodal function. In real world optimization problem, the objective functions would be unknown or so called "black box" function:

\begin{equation}
f(x) = - \sin(3x^2) - x^2 + 1.3x\label{eq:objective_fn}
\end{equation}

\begin{figure}[H]
    \centering
    \includegraphics[scale=0.5]{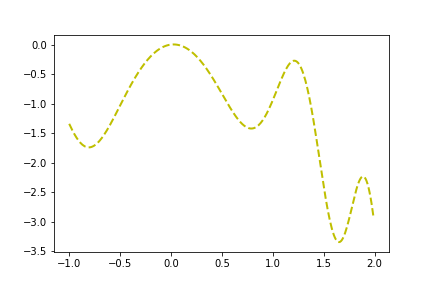}
    \caption{Objective function}
    \label{fig:objective_fn_1}
\end{figure}

This function has multiple local maximas and a global maxima one would intend to search via Bayesian optimization as shown in the Figure \ref{fig:objective_fn_1}

\begin{figure}[H]
    \centering
    \includegraphics[scale = 0.5]{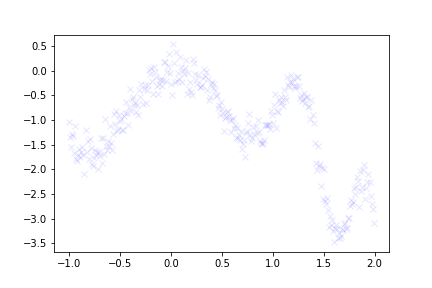}
    \caption{Noisy samples}
    \label{fig:noisy_samples}
\end{figure}

In a real world application, the objective function is never pure. There is always noise introduced by various external factors depending on the use case. For example, in parameter tuning, external noise can be slight changes in computation power or memory. Hence, a noise is introduced to the Equation \ref{eq:objective_fn} to obtain noisy samples for experimentation as shown in the Figure \ref{fig:noisy_samples} 

All the experiments were run on the samples obtained in Figure \ref{fig:noisy_samples}. Two initial samples with $x = -0.9$ and $x = 0.9$ were drawn for Bayesian optimization and used throughout the beginning of each experiment for consistency. The implementation is in Numpy, Matplotlib and Scipy. Scipy's interal libraries for the optimizers are used, however the implementation of acquisition functions is custom or in-house. The number of iterations were chosen to be 10 as the stopping conditition for Bayesian Optimization. It can be seen in the subsequent sub sections that the near-optimum for both acquisition functions was reached well before the stopping condition of Bayesian Optimization irrespective of the inner optimizer used. The following sub sections illustrate the various results obtained by using different optimizers on different acquisition functions.

\subsection{Limited-memory BFGS}
Limited-memory BFGS (L-BFGS or LM-BFGS) is an optimization algorithm in the family of quasi-Newton methods that approximates the Broyden–Fletcher–Goldfarb–Shanno algorithm (BFGS) using a limited amount of computer memory. It is a popular algorithm for parameter estimation in machine learning. The algorithm's target problem is to minimize $f(x)$ over unconstrained values of the real-vector $x$ where $f$ is a differentiable scalar function.

Like the original BFGS, L-BFGS uses an estimate of the inverse Hessian matrix to steer its search through variable space, but where BFGS stores a dense $n \times n$ approximation to the inverse Hessian ($n$ being the number of variables in the problem), L-BFGS stores only a few vectors that represent the approximation implicitly. \cite{lbfgs}

\subsubsection{Expected Improvement}
\begin{figure}[H]
    \centering
    \includegraphics[scale = 0.5]{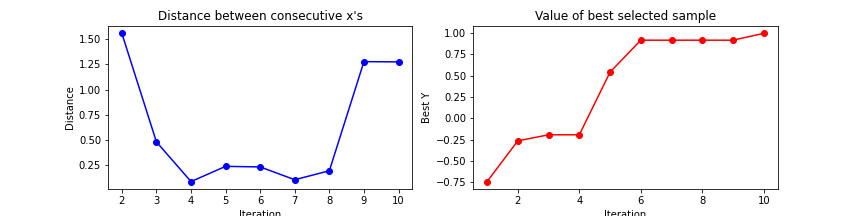}
    \caption{Convergence of Expected Improvement using L-BFGS}
    \label{fig:ei_lbfgs_convergence}
\end{figure}

Bayesian Optimization was run on the objective function \ref{eq:objective_fn} using Expected Improvement as acquisition function along with L-BFGS optimizer. The convergence is plotted in the Figure \ref{fig:ei_lbfgs_convergence}

\begin{figure}[H]
    \centering
    \includegraphics[scale=0.5]{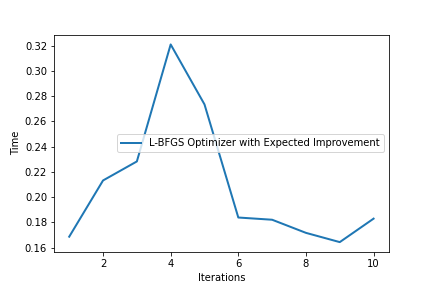}
    \caption{Time taken by Expected Improvement to propose new sample}
    \label{fig:ei_lbfgs_time_iterations}
\end{figure}

The time required to propose a new sample to be explored by Expected Improvement using L-BFGS was recorded. Figure \ref{fig:ei_lbfgs_time_iterations} shows the proposal time.

Closely evaluating Figure \ref{fig:ei_lbfgs_convergence} and \ref{fig:ei_lbfgs_time_iterations} we see that the time required by L-BFGS to propose a new sample location for Expected Improvement kept increasing in the initial stages of the search. After closing towards the near-optimum value of the objective function, the time taken by L-BFGS to propose a new sample also decreased.
 
\subsubsection{Maximum Probability of Improvement}
\begin{figure}[H]
    \centering
    \includegraphics[scale = 0.5]{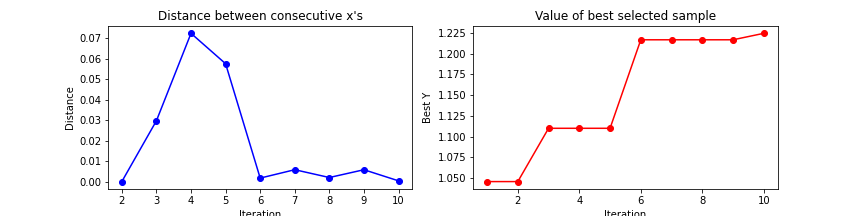}
    \caption{Convergence of Maximum Probability of Improvement using L-BFGS}
    \label{fig:mpi_lbfgs_convergence}
\end{figure}

Bayesian Optimization was then run using in-house implementation of Maximum Probability of Improvement along with L-BFGS optimizer. The convergence is plotted in the Figure \ref{fig:mpi_lbfgs_convergence}

\begin{figure}[H]
    \centering
    \includegraphics[scale=0.5]{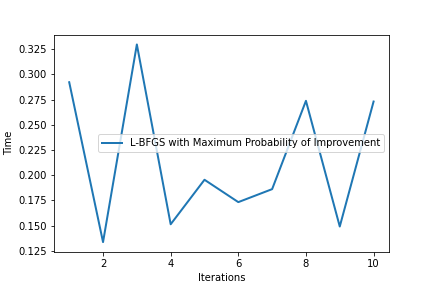}
    \caption{Time taken by Maximum Probability of Improvement to propose new sample}
    \label{fig:mpi_lbfgs_time_iterations}
\end{figure}

The time required to propose a new sample was recoreded similar to the previous experimentation. Figure \ref{fig:mpi_lbfgs_time_iterations} represents the proposal time.

Closely evaluating Figure \ref{fig:ei_lbfgs_convergence} and \ref{fig:ei_lbfgs_time_iterations} we see that the time required by L-BFGS to propose a new sample location for Maximum Probability of Improvement kept resonating between $0.125$ and $0.325$ seconds.
 
\subsection{Truncated Newton Optimizer}
Truncated Newton optimizer, also known as Hessian-free optimization, belongs to the family of optimization algorithms designed for optimizing non-linear functions with large numbers of independent variables. A truncated Newton method consists of repeated application of an iterative optimization algorithm to approximately solve Newton's equations, to determine an update to the function's parameters. The inner solver is truncated, i.e., run for only a limited number of iterations. It follows that, for truncated Newton methods to work, the inner solver needs to produce a good approximation in a finite number of iterations; conjugate gradient has been suggested and evaluated as a candidate inner loop. Another prerequisite is good preconditioning for the inner algorithm.
\subsubsection{Expected Improvement}
\begin{figure}[H]
    \centering
    \includegraphics[scale=0.5]{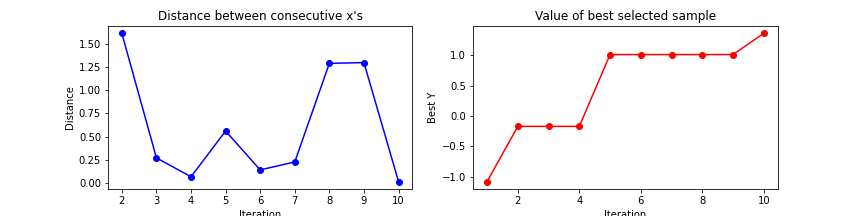}
    \caption{Convergence of Expected Improvement using TNC}
    \label{fig:ei_tnc_convergence}
\end{figure}

Experiments were repeated for TNC optimizer available in Scipy library, along with in-house implementation of Expected Improvement acquistion function. After the complete run of Bayesian Optimization with $10$ iterations, convergence of acquisition function was plotted as shown in Figure \ref{fig:ei_tnc_convergence}

\begin{figure}[H]
    \centering
    \includegraphics[scale=0.5]{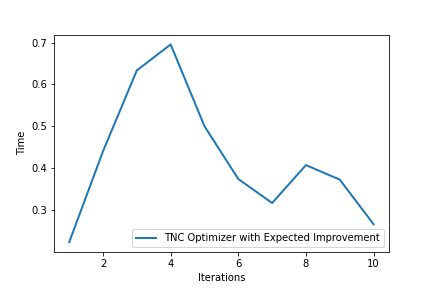}
    \caption{Time taken by Expected Improvement to propose new sample}
    \label{fig:ei_tnc_time_iterations}
\end{figure}

The time required for coming up with new sample point was recorded similar to previous optimizer for Expected Improvement. Figure \ref{fig:ei_tnc_time_iterations} shows the plot obtained.

Analysing Figure \ref{fig:ei_tnc_convergence} and \ref{fig:ei_tnc_time_iterations}, we can see that the time required by the optimizer to find the next sampling point kept increasing as Expected Improvement preferred exploration over exploitation which is evident from the Distance between the consecutive x's and the time taken.

\subsubsection{Maximum Probability of Improvement}

Similar experiments were carried out using Maximum Probability of Improvement and using TNC as the optimizer. Figure \ref{fig:mpi_tnc_convergence} shows the results obtained:

\begin{figure}[H]
    \centering
    \includegraphics[scale=0.5]{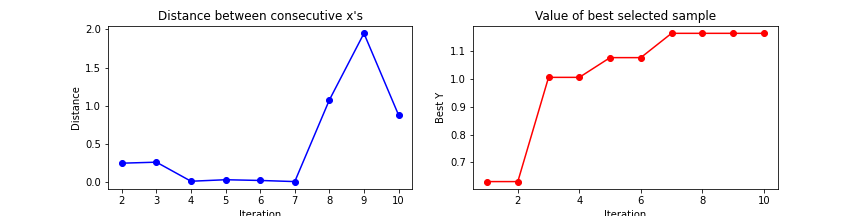}
    \caption{Convergence of Expected Improvement using TNC}
    \label{fig:mpi_tnc_convergence}
\end{figure}

\begin{figure}[H]
    \centering
    \includegraphics[scale=0.5]{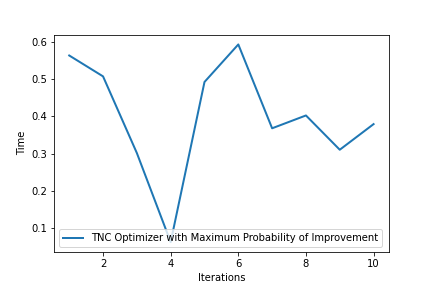}
    \caption{Time taken by Maximum Probability of Improvement to propose new sample}
    \label{fig:mpi_tnc_time_iterations}
\end{figure}

Upon plotting the time required for optimizing Maximum Probability of Improvement to obtain the next possible sample point, we get the variation as per Figure \ref{fig:mpi_tnc_time_iterations}

Analysis of Value of best selected sample in Figure \ref{fig:mpi_tnc_convergence} and Figure \ref{fig:mpi_tnc_time_iterations} shows an interesting trend. Whenever the value of the best selected sample remains the same in consecutive evaluations, we see a fall in time taken by the optimizer to suggest the next best sample.

\section{Conclusion}

\subsection{Maximum Probability of Improvement vs Expected Improvement}

There are many research works exploring the nature of Expected Improvement and Maximum Probability of Improvement concluding Expected Improvement is much efficient in finding the global optimum as compared to Maximum Probability of Improvement. However from the Figures \ref{fig:ei_tnc_convergence}, \ref{fig:ei_lbfgs_convergence}, \ref{fig:mpi_tnc_convergence} and \ref{fig:mpi_lbfgs_convergence} we see that MPI could reach closer to the global optimum much faster as compared to EI.

\begin{figure}[H]
    \begin{subfigure}{.5\textwidth}
        \centering
        \includegraphics[scale=0.5]{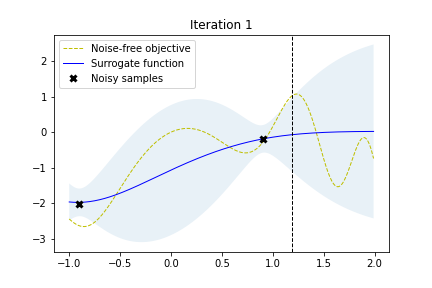}
        \caption{Gaussian Process with initial sampling}
        \label{fig:mpi_gp}
    \end{subfigure}%
    \begin{subfigure}{.5\textwidth}
        \centering
        \includegraphics[scale=0.5]{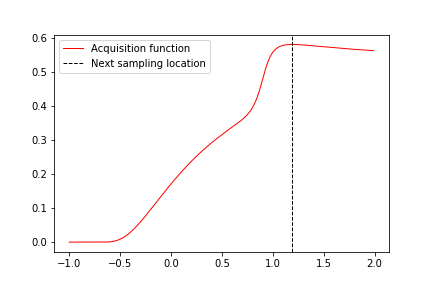}
        \caption{Acquisition function with next sampling location proposed}
        \label{fig:mpi_acquisition}
    \end{subfigure}
\caption{First iteration of Maximum Probability of Improvement suggesting sample location}
\label{fig:mpi_first_iteration}
\end{figure}

\begin{figure}[H]
    \begin{subfigure}{.5\textwidth}
        \centering
        \includegraphics[scale=0.5]{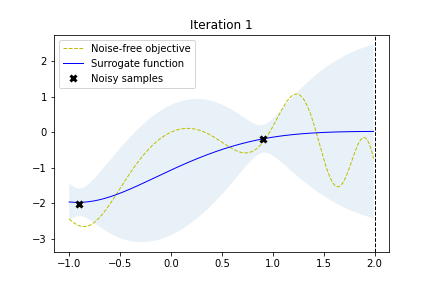}
        \caption{Gaussian Process with initial sampling}
        \label{fig:ei_gp}
    \end{subfigure}%
    \begin{subfigure}{.5\textwidth}
        \centering
        \includegraphics[scale=0.5]{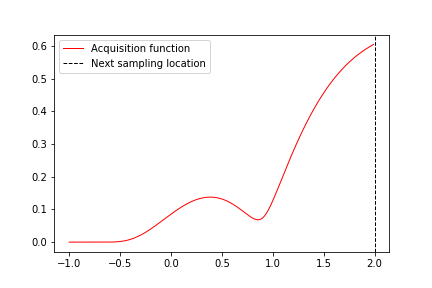}
        \caption{Acquisition function with next sampling location proposed}
        \label{fig:ei_acquisition}
    \end{subfigure}
\caption{First iteration of Expected Improvement suggesting sample location}
\label{fig:ei_first_iteration}
\end{figure}

Figure \ref{fig:mpi_first_iteration} shows the first iteration of Maximum Probability of Improvement acquisition function, where as Figure \ref{fig:ei_first_iteration} shows the first iteration of Expected Improvement acquisition function. From the experiments, we see that the initial points selected were close to the global maxima of the objective function. In such a scenario, Maximum Probability of Improvement suggests the next sampling point closer to the initial sample with greater value, thus preferring "Exploitation" as per the nature of MPI.

On the other hand, Expected Improvement, for the same sample points, prefer a sampling point away from both the initial samples, thus preferring "Exploration".
Hence, we see that Maximum Probability of Improvement is much efficient in finding the global optimum of the objective function in scenarios where the global optimum would be closer to one of the sample location and vice versa.

We conclude that the choice of the initial sampling plays an important role for both the acquisition functions.

\subsection{Limited-memory BFGS vs Truncated Newton optimizer}

\begin{table}[H]
\centering
\begin{tabular}{||c c c||} 
 \hline
  & MPI & EI \\ 
 \hline\hline
 L-BFGS & 0.215835 & 0.208968 \\ 
 TNC & 0.397903 & 0.423447\\ 
 \hline
\end{tabular}
\caption{Average time to suggest new sampling point in seconds}
\label{table:avg_time_optimization}
\end{table}

Table \ref{table:avg_time_optimization} shows the average time taken by L-BFGS and TNC optimizers on Maximum Probability of Improvement and Expected Improvement optimization functions for suggesting a new sample point over ten iterations.

We thus conclude that, L-BFGS optimizer works well on both acquisition functions and is a good fit for "inner optimization" problem.

\bibliographystyle{abbrv}
\bibliography{main}

\end{document}